\def\year{2021}\relax
\documentclass[letterpaper]{article}
\usepackage{aaai21}
\usepackage{times}
\usepackage{helvet}
\usepackage{courier}
\usepackage[hyphens]{url}
\usepackage{graphicx}
\usepackage{natbib}
\usepackage{caption}
\usepackage{float}
\usepackage{amsmath,amssymb,bm}
\usepackage{algorithm}
\usepackage{algorithmic}
\graphicspath{{figures/}}
\makeatletter
\gdef\copyright@on{}
\makeatother
\title{Binary Iterative Method for Non-targeted Adversarial Attack}
\author{
    Naman Goyal,\textsuperscript{\rm 1}
    Milan Chaudhari\textsuperscript{\rm 1}\\
}
\affiliations{
    \textsuperscript{\rm 1}Indian Institute of Technology, Ropar\\
}

\begin{document}
\maketitle

\begin{abstract}
Adversarial attacks have come a long way in guiding and providing additional training and test data for both adversarial training and adversarial robustness validations. In turn, the development of new adversarial attacks helps in improving reliability and sentiment values associated with classification, detection and various other areas of computer vision by exposing ‘piecewise linearity’ of deep learning-based models. Moreover, adversarial attacks and adversarial robustness are mathematically defined problems which can be optimised directly using end-to-end differentiable search techniques. Hence adversarial robustness is usually more widely applicable than the other types of robustness metrics including Corruption and Perturbation Robustness. Consequently, the generation of new kinds of adversarial attacks is beneficial for robustness testing. Generally, adversarial attacks can be classified as targeted and non-targeted methods based on the fact that the original image is modified to misclassify to a particular target class or any-incorrect class (non-targeted) for an unknown machine learning classifier. However, finding the optimized input data points and hyper-parameters for the generation of non-targeted adversarial attacks still remains a challenge in currently available methods like the Fast Gradient Method, Basic Iterative Method, Virtual Adversarial Method. To this end, we propose a new method “Binary Iterative Method” using divide-and-conquer paradigm to optimise parameters and hyper-parameters for the generation of non-targeted methods. Further, we compare our method to other gradient-based adversarial attacks which are evaluated over pre-trained networks like InceptionV3, InceptionV2, ResNet V2 152 over classification tasks. We observed that our “Binary Iterative Method” performed well over most of the other gradient-based methods on randomly-sampled 1000 images from the standard ImageNet dataset. In fact, in addition to outperforming all other methods,  our “Binary Iterative Method” qualitatively makes the classifier misclassify with very high confidence up to 0.995 while reducing the probability of true label to 2.21e-09 $(\approx 0)$.
\end{abstract}

\section{Introduction}

\noindent Adversarial examples refer to the transforming the input by an adversary so that any general machine learning model misclassify i.e the perturbed input results in the model outputting an incorrect answer with high confidence.

\begin{figure}
\includegraphics[width=\linewidth]{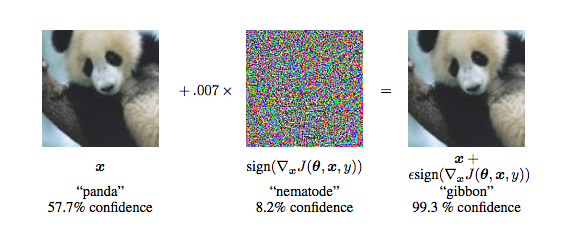}
\caption{
Explaining and Harnessing Adversarial Examples \cite{goodfellow2014explaining}
}

\label{fig:concept}
\end{figure}

The input is transformed by a small perturbation which looks like random noise, and can be so subtle that a human observer does not even notice the modification at all, yet the classifier still makes a mistake. See figure\ref{fig:concept}

There are two versions of the problem targeted and non-targeted both of which work on the generally unknown machine learning classifier. While in non-targeted attack aim is to just misclassify to any other class, in targeted one the aim is to misclassify to a particular class. Henceforth, the rest of the discussion is focused on non-targeted attack model.

The useful of new non-targeted adversarial attack methods in two folds. Firstly, generated adversarial examples are used to augment the training data, also known as adversarial training. Adversarial training is used to make the underlying machine learning model more adversarial robust.  Particularly, adversarial examples pose security concerns because they could be used to perform an attack on machine learning systems, even if the adversary has no access to the underlying model.

Secondly, new methods for generation of faster and better adversarial attacks are used to quantify the adversarial robustness of an underlying machine learning model.
Please not, the difference in the 2 use cases is that model never sees the generated adversarial examples in case 2 since it is only present in testing data. Apart form the usual performance measures, adversarial robustness itself becomes more useful parameter when choosing among different machine learning models. This is especially useful in an real-time setting like live object detection, object-tracking, person-identity verification etc.

While the noise added looks completely random; it is carefully computed using a another machine learning model various methods. Our main focus is on analyzing these methods and evaluating these methods on some state-of-the neural networks.

The main contributions if are our paper on 3 fronts.
\begin{enumerate}

    \item We suggest current gradient based adversarial attack generation methods cannot guarantee an optimal value of generated adversarial examples even after large number of epochs. This is due to the intrinsic limitations of the $ \epsilon -ball $ search techniques.

    \item To this end we contribute an non-targeted adversarial attack that very closely approximates finding multiple local minimas over finite number of epochs.

    \item Further, our method showcases much better results that current state-of-the-art methods for non-targeted adversarial attacks based on ImageNet dataset.
\end{enumerate}

\section{Related Work}

Adversarial attacks aim to fool deep learning methods by making them predict incorrectly for input that can easily classified by humans. There are several kinds of adversarial attacks ranging from gradient based attacks such as momentum iterative attack \cite{dong2018boosting}, FGSM, PGD \cite{madry2017towards}, distributionally adversarial attack \cite{zheng2019distributionally}, JSMA, DeepFool \cite{moosavi2016deepfool} and elastic-net attack \cite{chen2017ead}. Some attacks are based on adversarial patch and GAN-based attacks \cite{hu2017generating}.

\section{Preliminaries}
Our work is based on adversarial attacks on machine learning models. These attacks were first time introduced by \cite{gu2014towards} where they showed  deep neural networks (DNNs) to be highly susceptible to well-designed, small perturbations at the input layer.

While the early attempts at explaining this phenomenon focused on non-linearity and overfitting \cite{goodfellow2014explaining} argued instead that the primary cause of neural networks' vulnerability to adversarial perturbation is their linear nature i.e. piece-wise linearity among the various layers of neural networks. And came up with the idea of simple and fast method of generating adversarial examples.

The major solutions are now available through cleverhans \cite{papernot2016cleverhans} where the main focus of attack is based on the idea of how the weight updates of a neural network work.

The basic idea is to attack the way a neural network updates is parameter i.e. back-propagation.

\begin{equation}
w_{t+1} = w_{t} -\alpha {\nabla_{w_{t}}  Loss}
\end{equation}

To increase the $Loss$ we instead transform each input image $x$ by $\nabla_{x}  Loss$

\begin{equation}
x_{perturb} = x + eta =  x + \alpha \nabla_{x}  Loss
\end{equation}

then any neural network learned on the $x$ when tested on $x_{perturb}$ would decrease the original output confidence for true class of $x$.

\begin{figure*}
\includegraphics[width=\linewidth]{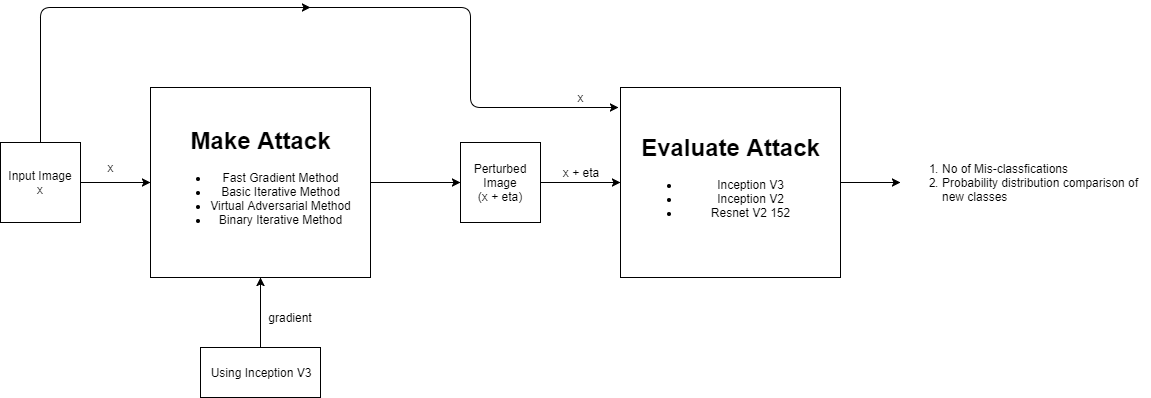}
\caption{
Method of transforming the image
}
\label{fig:method}
\end{figure*}
Some of the current methods which are used to compare our results are
\begin{enumerate}
\item \textbf{Fast Gradient Sign Method (FGSM)} In this method we take a machine learning model and an image, calculate gradients of the loss function(the cross-entropy error) w.r.t. input image, calculate the normalized gradients (In this case the sign of gradients) and add it to the input image after multiplying with a small number (epsilon: the step-size).

\begin{equation}
x_{perturb} = x + eps* sign ( {\nabla_{x} Loss} )
\end{equation}

\item \textbf{Basic Iterative Method (BIM)} This the application of the Fast Gradient Method iteratively over the  perturbed image by fixing a small step-size and iteratively update the image.
\begin{equation}
    x_{perturb}^{1} = x
\end{equation}
\begin{equation}
    x_{perturb}^{t+1} = x_{perturb}^{t} + eps\_iter* sign ( {\nabla_{x_{perturb}^{t}} Loss} )
\end{equation}

\item \textbf{Virtual Adversarial Method (VAM)} \cite{miyato2015distributional} This method is based on the careful update using KL loss based on logits (the model's unnormalized output or the input to the softmax layer)

\end{enumerate}
\section{Methodology}

Our approach \textbf{Binary Iterative Method (BinIM)} builds upon the Basic Iterative Method with the underlying approximation of using binary search to look for local minimas. Admittedly, binary search is used for  are searching in a concave region of  $\nabla_{x} Loss$ function.

We now need to find maximum of the same i.e. a point with zero gradient. Since the gradient in concave part of graph behaves like a array were values on the left side are positive and values on the left side are negative and they are in \textit{sorted order} we need to search the value 0.
\\

\begin{tabular}{|c|c|c|c|c|c|c|c|c|c|c|c|c|}
\hline
    200 & 100 & 50 & 20 & 0 & -19 & -50 & -100 & -224 \\
    \hline
\end{tabular}
\\

So we do a binary search where we start with a very high value of $eps\_iter^{1}= eps$ then each iteration reduce the $eps\_iter^{t+1} = eps\_iter^{t}/ 2$
i.e.

We further argue that, the current state-of-the-art methods including Fast  Gradient  Sign  Method (FGSM), Basic  Iterative  Method  (BIM), Virtual  Adversarial  Method  (VAM)  that use $ \epsilon -ball $ i.e. to search in an area intrinsically disadvantaged, since even after significant time spent on search, it's not guaranteed to be learn to any local minima.

\begin{figure}
\centering
\includegraphics[width=0.4\linewidth]{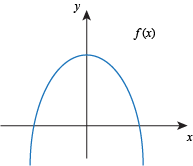}
\caption{Maximising the error using binary search to approximate the local minimas for a single round.}
\label{fig:concave}
\end{figure}

On the other hand, using a Binary Search Iterative Method, guarantees in the same search time ensures we are much closer to at least some local minima. Specifically, in a convex (concave) setting, binary search is guaranteed to be near global maxima (minima), while in other plains its guaranteed to be near at least some local maxmima (minima). See figure\ref{fig:concave}

Further we observe that, using a Binary Search Iterative Method to generate adversarial attack is much more useful when used with restart across multiple rounds. See algorithm\ref{alg:binim}

\begin{algorithm}
	\caption{BinIM algorithm (Non-targeted adversarial attack) repeated over multiple rounds}
	\label{alg:binim}
	\begin{algorithmic}
		\REQUIRE A known classifier $K$ to generate non-targeted adversarial attack for unknown classifier, $R$ number of rounds to re-initialize; permissible epochs per round; $E$ maximum permissible epochs per round;
		\STATE \textbf{Optimisation Objective}
		Find an optimal value of $eps$ starting with $eps\_iter^{1}$ using binary search until the generated attack using $ x_{perturb} $ is a stronger attack.
		\FOR {$r$ = $0$ to $R-1$}

		\STATE \textbf{Restart and Initialization} of the attack parameters as

		$ x_{perturb}^{1} = x$\\
        $eps\_iter^{1}= eps$

		\FOR {$e$ = $0$ to $E-1$}
		\STATE \textbf{Compute} the gradient for this iteration. Loss is w.r.t known classifier $K$

		$ x_{perturb_r}^{t+1} = x_{perturb_r}^{t} + eps\_iter_{r}^{t}* sign ( {\nabla_{x_{perturb_r}^{t}} Loss} ) $

		\STATE \textbf{Update} the $eps\_iter_{r}$ until binary search objective is maximised

        $ eps\_iter_{r}^{t+1}= eps\_iter_{r}^{t}/2$

		\ENDFOR
		\ENDFOR
		\STATE \textbf{return} the best $ eps\_iter^{E}_{r} $ across rounds. This is an approximation to  choosing the best local minima among multiple local minimas.

	\end{algorithmic}

\end{algorithm}

Update region
\begin{equation}
\label{eq:update}
\begin{cases}
& x_{perturb}^{1} = x \\
& eps\_iter^{1}= eps
\end{cases}
\end{equation}

Update as
\begin{equation}
\begin{cases}
x_{perturb}^{t+1} = x_{perturb}^{t} + eps\_iter^{t}* sign ( {\nabla_{x_{perturb}^{t}} Loss} )\\
eps\_iter^{t+1}= eps\_iter^{t}/2
   \end{cases}
   \label{eqn:}
\end{equation}

\section{Experiment}

The aim of adversarial attack is to reduce the accuracy of classification.
The approach to generate result is as described in figure\ref{fig:method} i.e.
\begin{enumerate}
\item Sample 1000  images  from  the  standard  ImageNet dataset.
\item Transform the images using 3 available methods i.e.  Fast Gradient Method, Basic Iterative Method, Virtual Adversarial Method and our method i.e.  Binary Iterative Method using gradient of loss computed over Inception V3 network.
\item To validate our results we then evaluate the accuracy of classification over the original data set and 4 perturbed outputs over 3 different state-of-the networks for classification i.e. InceptionV3, InceptionV2, ResNet V2 152.

\end{enumerate}

\begin{table}
  \centering
    \begin{tabular}{|c|c|}
\hline
\textbf{Model} & \textbf{Clean Accuracy} \\
\hline
     Inception\_V3 & 0.958 \\
     \hline
     Inception\_V2 & 0.855 \\
     \hline
     Resnet\_V2\_152 & 0.906 \\
     \hline
\end{tabular}

\caption{Clean test-accuracy over the original dataset}
\end{table}

\begin{table}
  \centering
\begin{tabular}{|c|c|c|c|c|}
\hline
     \textbf{Model} & \textbf{FGSM} & \textbf{BIM} & \textbf{VAM} & \textbf{BinIM (Ours)} \\
\hline
     \textbf{Inception\_V3} & 0.356 & 0.0252 & 0.628 & 0.009 \\
\hline
     \textbf{Inception\_V2} & 0.882 & 0.839 & - & 0.683 \\
\hline
     \textbf{Resnet\_V2\_152} & 0.835 & 0.828 & - & 0.668 \\
\hline

\end{tabular}
\caption{Adversarial testing Accuracy over the modified dataset}
\label{table:results}
\end{table}

\noindent
FGSM = Fast Gradient Sign Method\\
   BIM = Basic Iterative Method\\
    VAM = Virtual Adversarial Method\\
     BinIM = Binary Iterative Method \textit{(our method)}


\begin{table}[H]

  \begin{center}

\begin{tabular}{c c}
\textbf{Original Image} & \textbf{Modified Image using BinIM} \\
\includegraphics[scale=0.4]{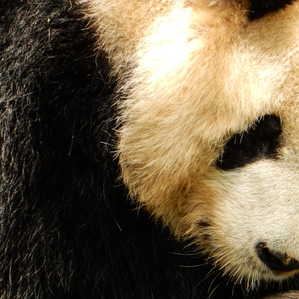} & \includegraphics[scale=0.4]{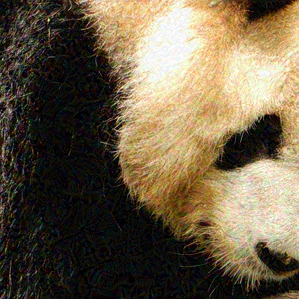} \\
\includegraphics[scale=0.4]{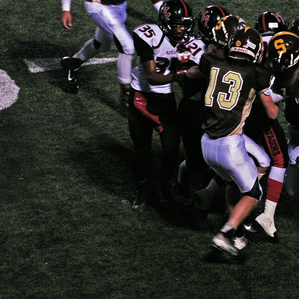} & \includegraphics[scale=0.4]{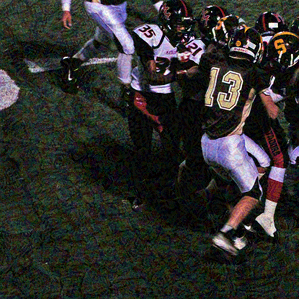} \\
\includegraphics[scale=0.4]{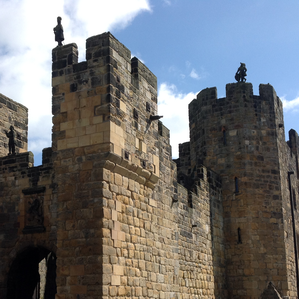} & \includegraphics[scale=0.4]{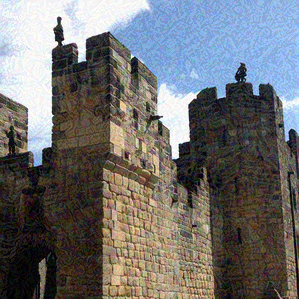} \\
\includegraphics[scale=0.4]{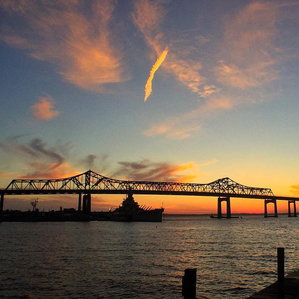} & \includegraphics[scale=0.4]{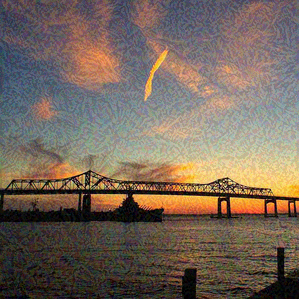} \\
\end{tabular}

  \end{center}
\caption{Sample outputs generated using Binary Iterative Method}
\label{table:images}
\end{table}

\begin{table*}
\begin{center}

\begin{tabular}{|c|c|c|c|c|c|}
\hline
    Initial Predicted label & Initial Prob & Modified Label & Modified Prob & Modified Prob for initial label & True Label \\
\hline
    306 & 0.807618 & 303 & 0.9995 & 2.21524e-09 & 306 \\
\hline
    884 & 0.995941 & 873 & 1.0 & 6.9128e-17 & 884 \\
\hline
    244 & 0.946103 & 152 & 1.0 & 2.88032e-16 & 244 \\
\hline
    560 & 0.834364 & 542 & 1.0 & 6.43702e-20 & 560 \\
\hline
    439 & 0.542039 & 371 & 0.989378 & 3.6594e-08 & 439 \\
\hline
    991 & 0.832426 & 954 & 1.0 & 1.81221e-19 & 991 \\
\hline
    950 & 0.97152 & 114 & 1.0 & 7.20899e-17 & 950 \\
\hline
    854 & 0.988273 & 778 & 1.0 & 1.17832e-13 & 854 \\
\hline
    610 & 0.814898 & 657 & 0.999993 & 4.58456e-12 & 610 \\
\hline
    610 & 0.434363 & 437 & 0.999994 & 2.39441e-11 & 610 \\
\hline
    583 & 0.337361 & 455 & 0.999609 & 2.3094e-22 & 583 \\
\hline
\end{tabular}
\caption{Sample Transformation of Probabilities using Binary Iterative Method (Ours) evaluated on Inception\_V3}
\label{table:trans}
\end{center}
\end{table*}

\section{Discussions}

Based on results presented in Table\ref{table:results} and images presented in Table\ref{table:images} we observe the following trends

\paragraph{Observation} The accuracy over Inception\_V3 decreases the most over other models to evaluate the attack.
\paragraph{Explanation} The gradient to make the attack was using Inception\_V3 hence it is expected to make maximum classifications over the Inception\_V3 itself.

\paragraph{Observation} All the four methods are not much effective over the Resnet\_V2 and Inception\_V2, but still they are able to reduce confidence of true label.
\paragraph{Explanation}``An intriguing aspect of adversarial examples is that an example generated for one model is often misclassified by other models, even when they have different architecures or were trained on disjoint training sets. Moreover, when these different models misclassify an adversarial example, they often agree with each other on its class.   Explanations based on extreme non-linearity and over-
fitting cannot readily account for this behavior. This behavior is especially surprising from the view of the hypothesis that adversarial examples finely tile space like the rational numbers among the reals, because in this view adversarial examples are common but
occur only at very precise locations.", as stated by  ~\cite{goodfellow2014explaining}.

\paragraph{Observation} The Binary Iterative Method (BinIM) causes the most number of misclassifciations.
\paragraph{Explanation} The Binary Iterative Method (BinIM) is better than the combination of techniques used in the FGSM, BIM hence is expected to give better result i.e. more misclassfications.

\paragraph{Observation} Few the modified images are indistinguishable from the original one such as images 1 and 2; while the 3rd and 4th images are distinguishable.

\paragraph{Explanation} This depends on the spatial features present in the images such as in both the 3rd and 4th images; the presence of sky causes a human observer to notice the change in the color intensity and the change in color intensity.

\paragraph{Observation}

The table \ref{table:trans} represents how the model 'Inception\_V3' were initially classifying and how the classification changes after modification in the input images.

Consider the first example, initially the model predicted the label 306 with probability of 0.807, which is the true label.

While after modification in the image, the model predicts the label 303, which not only is wrong label, but model's is confident with probability is 0.9995.
Moreover the probability of the true label has reduced to 2.21e-09 $(\approx 0)$, which is quite low.

\paragraph{Explanation} This shows the proposed Binary Iterative Method is able to generate a good adversarial examples and able to reduce the confidence of the original class to great extent.

\section{Conclusions and Future Works}

The proposed Binary Iterative Methods is able to generate adversarial examples which reduce the confidence of the original label to a great extent which was further validated using results from different state-of-the neural network over 1000 image data set sampled from Image-Net. Our BinIM method can be used for adversarial training to reduce the impact of other Adversarial Attack. Our main contributions hence includes (a) argument on failure of current state-of-the-art adversarial attack generation methods due to the intrinsic limitations of the $ \epsilon -ball $ search techniques. (b) proposal of a new Binary Iterative Method (BinIM) inspired from Binary Search for non-targeted adversarial attack that closely approximates finding multiple local minimas over finite number of epochs. (c) our method showcases much better results that current state-of-the-art methods for non-targeted adversarial attacks.

In future, since the community is always involving with new methods and which can be contrasted against this. One can further develop a more generic attack method which preforms well on various kinds of architectures. Additionally, one can extrapolate the Binary Iterative Method meant for non-targeted adversarial attack, for the targeted adversarial attack.

\bibliography{ref}

\begin{thebibliography}{10}
\providecommand{\natexlab}[1]{#1}
\providecommand{\url}[1]{\texttt{#1}}
\providecommand{\urlprefix}{URL }
\expandafter\ifx\csname urlstyle\endcsname\relax
  \providecommand{\doi}[1]{doi:\discretionary{}{}{}#1}\else
  \providecommand{\doi}{doi:\discretionary{}{}{}\begingroup
  \urlstyle{rm}\Url}\fi

\bibitem[{Chen et~al.(2017)Chen, Sharma, Zhang, Yi, and Hsieh}]{chen2017ead}
Chen, P.-Y.; Sharma, Y.; Zhang, H.; Yi, J.; and Hsieh, C.-J. 2017.
\newblock Ead: elastic-net attacks to deep neural networks via adversarial
  examples.
\newblock \emph{arXiv preprint arXiv:1709.04114} .

\bibitem[{Dong et~al.(2018)Dong, Liao, Pang, Su, Zhu, Hu, and
  Li}]{dong2018boosting}
Dong, Y.; Liao, F.; Pang, T.; Su, H.; Zhu, J.; Hu, X.; and Li, J. 2018.
\newblock Boosting adversarial attacks with momentum.
\newblock In \emph{Proceedings of the IEEE conference on computer vision and
  pattern recognition}, 9185--9193.

\bibitem[{Goodfellow, Shlens, and Szegedy(2014)}]{goodfellow2014explaining}
Goodfellow, I.~J.; Shlens, J.; and Szegedy, C. 2014.
\newblock Explaining and harnessing adversarial examples.
\newblock \emph{arXiv preprint arXiv:1412.6572} .

\bibitem[{Gu and Rigazio(2014)}]{gu2014towards}
Gu, S.; and Rigazio, L. 2014.
\newblock Towards deep neural network architectures robust to adversarial
  examples.
\newblock \emph{arXiv preprint arXiv:1412.5068} .

\bibitem[{Hu and Tan(2017)}]{hu2017generating}
Hu, W.; and Tan, Y. 2017.
\newblock Generating adversarial malware examples for black-box attacks based
  on gan.
\newblock \emph{arXiv preprint arXiv:1702.05983} .

\bibitem[{Madry et~al.(2017)Madry, Makelov, Schmidt, Tsipras, and
  Vladu}]{madry2017towards}
Madry, A.; Makelov, A.; Schmidt, L.; Tsipras, D.; and Vladu, A. 2017.
\newblock Towards deep learning models resistant to adversarial attacks.
\newblock \emph{arXiv preprint arXiv:1706.06083} .

\bibitem[{Miyato et~al.(2015)Miyato, Maeda, Koyama, Nakae, and
  Ishii}]{miyato2015distributional}
Miyato, T.; Maeda, S.-i.; Koyama, M.; Nakae, K.; and Ishii, S. 2015.
\newblock Distributional smoothing with virtual adversarial training.
\newblock \emph{arXiv preprint arXiv:1507.00677} .

\bibitem[{Moosavi-Dezfooli, Fawzi, and Frossard(2016)}]{moosavi2016deepfool}
Moosavi-Dezfooli, S.-M.; Fawzi, A.; and Frossard, P. 2016.
\newblock Deepfool: a simple and accurate method to fool deep neural networks.
\newblock In \emph{Proceedings of the IEEE conference on computer vision and
  pattern recognition}, 2574--2582.

\bibitem[{Papernot et~al.(2016)Papernot, Goodfellow, Sheatsley, Feinman, and
  McDaniel}]{papernot2016cleverhans}
Papernot, N.; Goodfellow, I.; Sheatsley, R.; Feinman, R.; and McDaniel, P.
  2016.
\newblock cleverhans v1.0.0: an adversarial machine learning library.
\newblock \emph{arXiv preprint arXiv:1610.00768} .

\bibitem[{Zheng, Chen, and Ren(2019)}]{zheng2019distributionally}
Zheng, T.; Chen, C.; and Ren, K. 2019.
\newblock Distributionally adversarial attack.
\newblock In \emph{Proceedings of the AAAI Conference on Artificial
  Intelligence}, volume~33, 2253--2260.

\end{thebibliography}

\appendix

\section{Appendix: Comparison $ \epsilon -ball $ and Binary Search}

\begin{figure}
\centering
\includegraphics[width=0.7\linewidth]{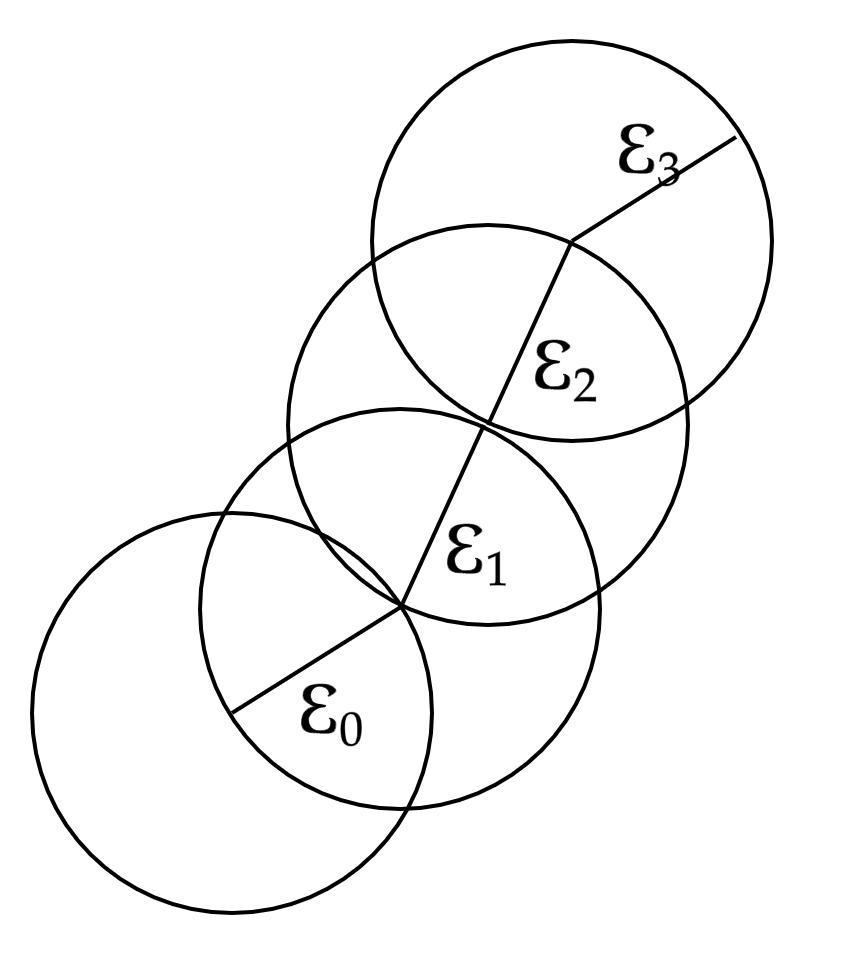}
\caption{Searching using $ \epsilon -ball $ search. As one moves across the $ \epsilon_{0} \rightarrow  \epsilon_{1} \rightarrow \epsilon_{2} \rightarrow \epsilon_{3} $ we can't be sure of the landing on a minina (maxima) since $ \epsilon $-search is mathematically not defined to optimise in a given an unbounded search space i.e. the radius of search increases as shown above.}
\label{fig:eps_search}
\end{figure}

The objective function for generation for adversarial attack is given by equation~\ref{eq:sum} wherein we are optimising for the mis-classification of an unknown classifier  using an approximation i.e. $ eps* sign ( {\nabla_{x} Loss})$ from a known classifier.

\begin{equation}
x_{perturb} = x + eps* sign ( {\nabla_{x} Loss} )
\label{eq:sum}
\end{equation}

Further, there are 2 types of optimisation possible -
\begin{enumerate}
    \item Using a different architecture which in turns changes the $sign ( {\nabla_{x} Loss})$.
    \item Using a different local-search technique for $eps$ in the search space defined by $sign ( {\nabla_{x} Loss})$. The  $sign ( {\nabla_{x} Loss})$ in turns gets updated in each iteration based on defined $eps$ in last iteration as per equation~\ref{eqn:iterative}
\end{enumerate}

We argue that the current state-of-the-art methods including Fast  Gradient  Sign  Method (FGSM), Basic  Iterative  Method  (BIM), Virtual  Adversarial  Method  (VAM)  that use $ \epsilon -ball $ search are intrinsically disadvantaged, since even after significant time spent on search, the returned values are not guaranteed to be close to any local minima. Moreover, the
$ \epsilon -ball $ search is not meant for optimisation in an unbounded setting i.e. radius of search increases if the next found value is out of the current radius as shown in figure ~\ref{fig:eps_search}.

\begin{equation}
    \begin{cases}
    x_{perturb}^{t+1} = x_{perturb}^{t} + eps\_iter^{t}* sign ( {\nabla_{x_{perturb}^{t}} Loss} )\\
    eps\_iter^{t+1}= eps\_iter^{t}/2
   \end{cases}
   \label{eqn:iterative}
\end{equation}

As shown in figure~\ref{fig:eps_search} the $ \epsilon -ball $ methods don't take advantage of properties of search space and much research efforts has been put forward towards optimising the $sign ( {\nabla_{x} Loss})$ through different architectures.

In contrast, our work focuses on using a better method of local search to find the optimal value for a given architecture. For this we compare the tradition $ \epsilon -ball $ search which is $O(n)$ in time complexity with the proposed application of Binary-Search i.e. $O(log n)$ for finding optimal value.

\begin{figure}
\centering
\includegraphics[width=1.0\linewidth]{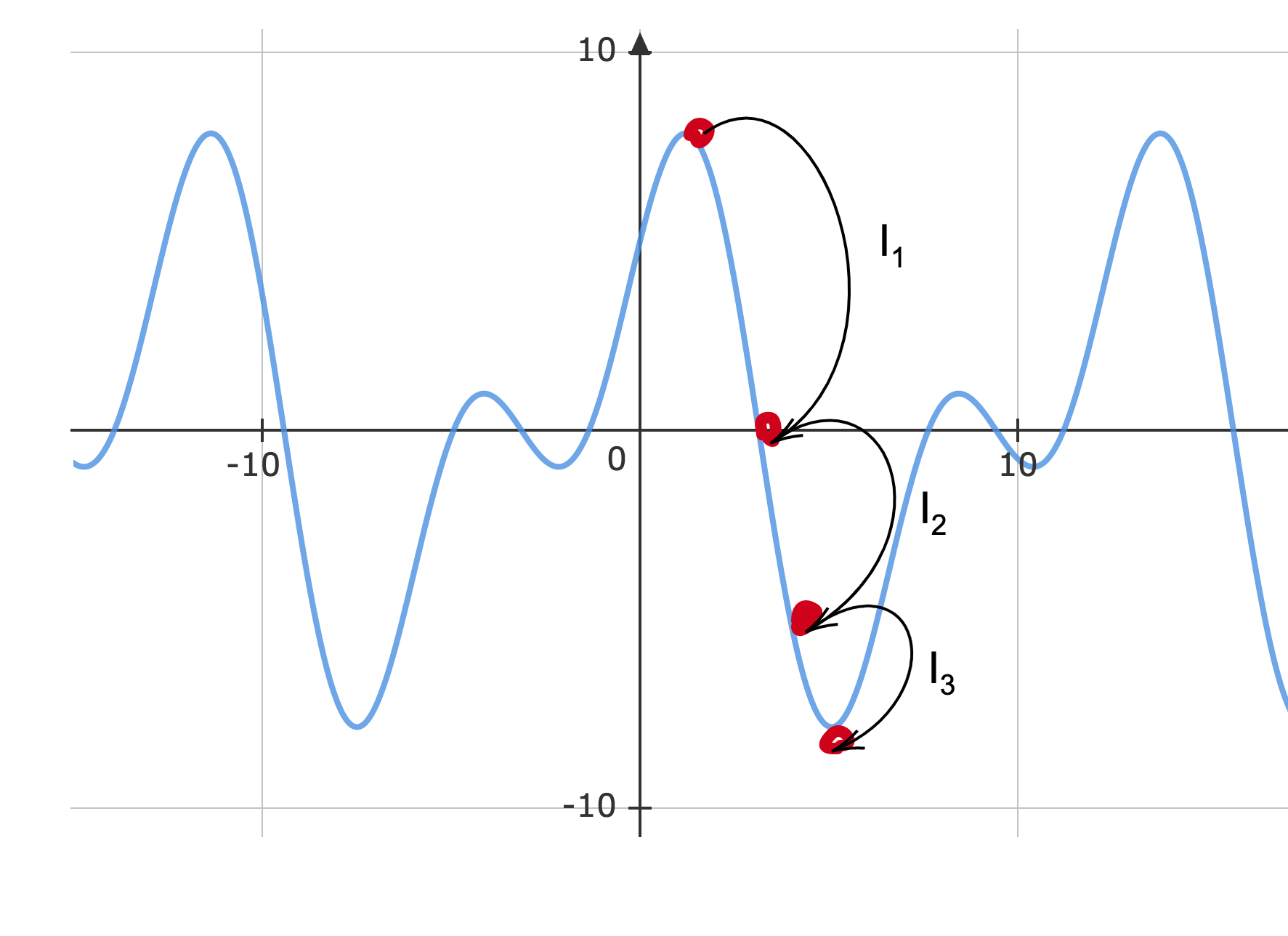}
\caption{Searching using Binary Search. As one moves across the iterations $ I_{1} \rightarrow  I_{2} \rightarrow I_{3} $, binary search ensures we reach closer to local minima (maxima) in any given objective function.}
\label{fig:binary_search}
\end{figure}

For instance, in figure~\ref{fig:binary_search} in 3 iterations we land in local minima, since in binary search we reduce the search space by half (or by a fixed constant every time). Even if we over shoot, we are guaranteed to settle \textit{near} on one of the local minima. For representation purpose the graph is a continuous  plot, but it can any general non-continuous non-differentiable function. This in turn performs better than all state-of-the-art methods as shown in the main paper.

In fact, using our Binary Search Iterative Method ensures that returned value for the adversarial attack objective is much closer to at least some local minima in the same (or less) number of epochs than traditional $ \epsilon -ball $ search. Specifically, we leverage the property of binary search that in a convex (concave) setting, binary search is guaranteed to be near global maxima (minima), while in other functions its guaranteed to be near at least some local maxmima (minima).

\end{document}